\pdfoutput=1

\documentclass[11pt]{article}

\usepackage[final]{ACL2023}

\usepackage{times}
\usepackage{latexsym}
\usepackage{bm}
\usepackage{hyperref}
\usepackage{color,colortbl}
\usepackage{stfloats}
\usepackage{enumitem}
\usepackage{booktabs}
\usepackage{nccmath}
\usepackage{multirow}
\usepackage{subfigure}
\usepackage{latexsym}
\usepackage{todonotes}
\usepackage{makecell}
\usepackage{dashrule}
\usepackage{amssymb}
\usepackage{bbding}

\usepackage{graphicx}

\definecolor{LightGray}{gray}{0.9}
\usepackage{soul}

\usepackage{arydshln}

\usepackage{amssymb}
\usepackage{pifont}
\newcommand{\cmark}{\ding{51}}%
\newcommand{\xmark}{\ding{55}}%

\usepackage[T1]{fontenc}

\usepackage[utf8]{inputenc}

\usepackage{microtype}

\usepackage{inconsolata}

\title{Multilingual Multi-Figurative Language Detection}

\author{
Huiyuan Lai, Antonio Toral, Malvina Nissim\\
CLCG, University of Groningen / The Netherlands\\
\texttt{\{h.lai, a.toral.ruiz, m.nissim\}@rug.nl}
}

\begin{document}
\maketitle
\begin{abstract}
Figures of speech help people express abstract concepts and evoke stronger emotions than literal expressions, thereby making texts  more creative and engaging. Due to its pervasive and fundamental character, figurative language understanding has been addressed in Natural Language Processing, but it's highly understudied in a multilingual setting and when considering more than one figure of speech at the same time.
To bridge this gap,
we introduce \textit{multilingual multi-figurative language modelling}, and provide a benchmark for sentence-level figurative language detection, covering three common figures of speech and seven languages. 
Specifically, we develop a framework for figurative language detection based on template-based prompt learning. In so doing, we unify multiple detection tasks that are interrelated across multiple figures of speech and languages, without requiring task- or language-specific modules. 
Experimental results show that our framework outperforms several strong baselines and may serve as a blueprint for the joint modelling of other interrelated tasks.
\end{abstract}

\section{Introduction}

Figurative language is ubiquitous in human language, allows us to convey abstract concepts and emotions, and has been embedded intimately in our cultures and behaviors~\citep{richard-etal-1994, Harmon2015FIGURE8AN}. 
In the hyperbolic sentence ``\emph{My heart failed a few times while waiting for the result.}'', the expression ``\emph{my heart failed a few times}''  
is not a literal heart-stop, it exaggerates the mood of when waiting for a possibly important result, thereby vividly showing anxiety. 

Recent years have seen a lot of interest in figurative language processing in the NLP community, including the successful organization of dedicated workshops~\citep{ws-2018-figurative, fig-lang-2020-figurative, flp-2022-figurative}. There are many works focusing on figurative language detection, mostly in English, including hyperbole~\citep{troiano-etal-2018-computational}, metonymy~\citep{nissim-markert-2003-syntactic}, metaphor~\citep{tsvetkov-etal-2014-metaphor}, idiom~\citep{liu-hwa-2018-heuristically} and sarcasm~\citep{hazarika-etal-2018-cascade}. In addition, researchers have started start to pay attention to figurative language detection in a multilingual scenario~\citep{tsvetkov-etal-2014-metaphor, tedeschi-etal-2022-id10m, aghazadeh-etal-2022-metaphors, tayyar-madabushi-etal-2022-semeval}, where models can exploit cross-lingual knowledge transfer~\citep{alexis2019cross}. Nonetheless, detection tasks for different figures of speech are usually studied independently of each other, which leads to having to train separate models for each figure of speech. However, different figures of speech are often related to each other, and therefore models can thus potentially benefit from cross-figurative knowledge transfer, as empirically shown by~\citet{lai-nissim-2022-multi} in a monolingual setting for English.

\begin{figure}[!t]
\centering
\includegraphics[scale=.32]{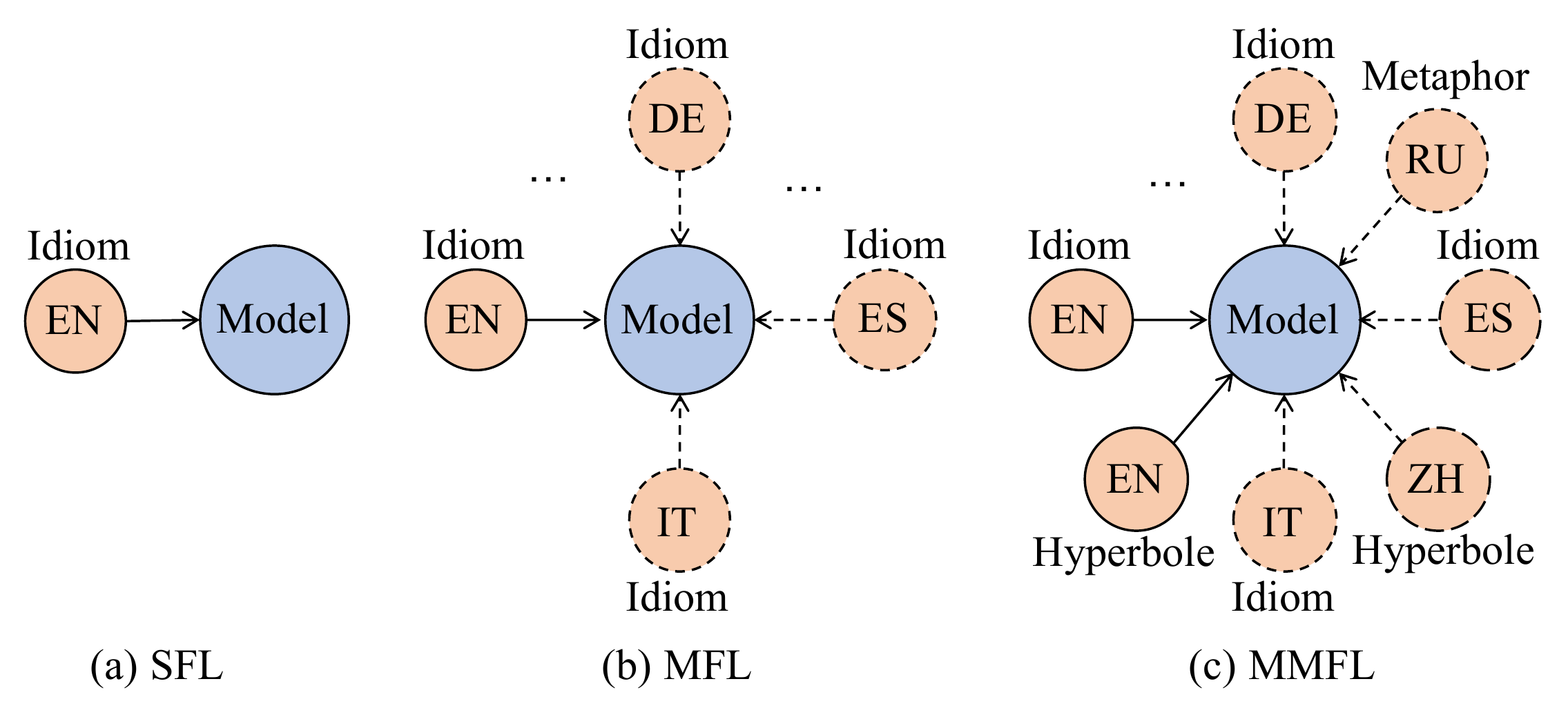}
\caption{Overview of three different modelling scenarios: (a) single figurative language (SFL), (b) multilingual figurative language (MFL) and (c) multilingual multi-figurative language (MMFL).}
\label{fig:overview}
\end{figure}

\begin{table*}[t]
\resizebox{\linewidth}{!}{%
\centering
\begin{tabular}{llcccc}
\toprule
\textbf{Reference} & \textbf{Description} & \textbf{M-Lang} & \textbf{M-Fig} & \textbf{M-Task} & \textbf{Level} \\
\hline
\citet{troiano-etal-2018-computational} & Hyperbole detection in English & \color{red}{\xmark} & \color{red}{\xmark} & \color{red}{\xmark} & Sentence\\
\citet{tedeschi-etal-2022-id10m} & Multilingual idiom detection & \color{blue}{\cmark} & \color{red}{\xmark} & \color{red}{\xmark} & Word\\
\citet{tayyar-madabushi-etal-2022-semeval} & Multilingual idiom detection & \color{blue}{\cmark} & \color{red}{\xmark} & \color{red}{\xmark} & Sentence\\
\citet{aghazadeh-etal-2022-metaphors} & Multilingual metaphor detection & \color{blue}{\cmark} & \color{red}{\xmark} & \color{red}{\xmark} & Sentence\\
\citet{lai-nissim-2022-multi} & Multi-figurative language generation in English & \color{red}{\xmark} & \color{blue}{\cmark} & \color{blue}{\cmark} & Sentence\\
\hdashline 
Our work & Multilingual multi-figurative language detection & \color{blue}{\cmark} & \color{blue}{\cmark} & \color{blue}{\cmark} & Sentence\\
\bottomrule
\end{tabular}}
\caption{\label{tab:comparisons}
A comparison of different previous works and the present one, according to whether they perform word-/sentence-level detection in a multilingual (M-Lang), multi-figurative (M-Fig), or multi-task (M-Task) fashion.
}
\end{table*}

In this paper we investigate how these related detection tasks can be connected and modelled jointly in a multilingual way (see Table~\ref{tab:comparisons}). 
To do so, we propose a multitask framework to model multilingual multi-figurative language detection at the sentence level. As shown in Figure~\ref{fig:overview}, our goal is to connect the detection tasks from different languages and different figures of speech, resulting in a unified model which can benefit from cross-lingual and cross-figurative knowledge transfer. 

Generally, a multi-task framework consists of shared modules and task-specific modules. With the development of pre-trained language models (PLMs), prompt learning offers the opportunity to model multiple tasks in a framework that does not require task-dependent parameters~\citep{radford2019language, brown2020, fu2022polyglot, mishra-etal-2022-cross}. With this method, task-specific language instructions are predefined and used to guide the model to handle different tasks.

In practice, we first formalize the figurative language detection task as a text-to-text generation problem, where the input is the source sentence while the target is a textual label (e.g.\ ``literal'' or ``idiomatic''). This method thus enables us to train our models in a sequence-to-sequence (seq2seq) fashion. We then prepend the prompt template to source sentences from various tasks when feeding them into the model. This connects multiple figures of speech and languages in a unified framework, also leading to a better understanding of how to jointly model tasks related to each other. 
We perform extensive experiments on three figures of speech: hyperbole, idiom, and metaphor, involving seven languages (English:EN, Chinese:ZH, German:DE, Spanish:ES, Italian:IT, Farsi:FA, and Russian:RU). 

Our main contributions are as follows: (i) We introduce the novel task of multilingual multi-figurative language detection, wherein we explore the potential of joint modelling. (ii) We introduce a multitask and multilingual framework based on 
prompt learning, which unifies interrelated detection tasks without task- nor language-specific modules. (iii) We evaluate the model's generalization capabilities across a range of figures of speech and languages: 
extensive experiments are run for different settings, including in-language, zero-shot, cross-lingual, and cross-figurative to show how the unified framework performs. (iv) Our framework may serve as a blueprint for joint modelling of other interrelated tasks, such as the detection of hate speech~\citep{waseem-hovy-2016-hateful}, offensive and abusive language~\citep{caselli-etal-2020-feel}, toxicity~\citep{pavlopoulos-etal-2020-toxicity}, as well as fake news and AI-generated content~\citep{zellers2019defending}. We have released our code and all preprocessed dataset.\footnote{ \url{https://github.com/laihuiyuan/MMFLD}}

\section{Related Work}

We briefly introduce the background of figurative language detection, from feature engineering to neural-based approaches, as well as prompt-based learning with PLMs. 

\subsection{Figurative Language Detection}


This task often involves word-level and sentence-level detection. Word-level detection is concerned with identifying the exact words within the context of a sentence used with a figurative meaning. Sentence-level detection, as a binary classification problem, requires to automatically detect whether a given sentence is literal or not. 

\paragraph{Feature Engineering}
Traditionally, 
researchers have investigated hand-engineered features to understand figurative usages. 
These features are primarily concerned with linguistic aspects, including word imageability~\citep{broadwell2013, troiano-etal-2018-computational}, word unexpectedness~\citep{troiano-etal-2018-computational}, syntactic head-modifier relations~\citep{nissim-markert-2003-syntactic}, abstractness and semantic supersenses~\citep{tsvetkov-etal-2014-metaphor}, property norms~\citep{bulat-etal-2017-modelling}, pragmatic phenomena~\citep{karoui-etal-2017-exploring}, together with other aspects such as sentiment~\citep{troiano-etal-2018-computational, rohanian-etal-2018-wlv}
and sensoriality~\citep{tekiroglu-etal-2015-exploring}. These features rely heavily on manual extraction and are very much task-dependent. Exploiting verb and noun clustering~\citep{shutova-etal-2010-metaphor} and bag-of-words approaches~\citep{koper-schulte-im-walde-2016-distinguishing} are common automated methods to reduce manual work.

\paragraph{Neural-Based Approaches} 
In the last decade, researchers have moved from feature engineering to neural-based modelling, using LSTM- 
\citep{wu-etal-2018-neural, gao-etal-2018-neural, mao-etal-2019-end, kong-etal-2020-identifying} and CNN-based approaches~\citep{wu-etal-2018-neural, kong-etal-2020-identifying} for figurative language detection. Most recently, PLMs have been used for this task, usually yielding new state-of-the-art results~\citep{su-etal-2020-deepmet, choi-etal-2021-melbert, zeng-bhat-2021-idiomatic, tedeschi-etal-2022-id10m}. Similar to other NLP tasks, researchers have also moved towards multilingual detection~\citep{tsvetkov-etal-2014-metaphor, tedeschi-etal-2022-id10m, tayyar-madabushi-etal-2022-semeval, aghazadeh-etal-2022-metaphors}, especially thanks to cross-lingual knowledge transfer via multilingual PLMs. All these works focus on single figures of speech, i.e.\ detecting whether a sentence (or each word in a sentence) contains a given figure of speech or it is literal. 
We take here the first step towards multilingual multi-figurative language modelling to introduce a unified framework for multiple languages and multiple figures of speech, focusing on sentence-level detection.

\subsection{Pre-Training and Prompt Learning}

Over the past few years, PLMs have brought NLP to a new era~\citep{devlin-etal-2019-bert, radford2018language, radford2019language, brown2020}. PLMs are pre-trained on massive textual data in a self-supervised manner, and then fine-tuned on downstream tasks with task-specific training objectives.
This paradigm, however, has to be adapted to different target tasks, where the task-specific objectives are different from the pre-training one, and the introduction of additional parameters such as a PLM-based classifier is at times necessary.

Prompt learning, a new learning paradigm based on PLMs, aims to make better use of pre-trained knowledge by reformulating tasks to be close to the pre-training objectives~\citep{liu2022pre}. Specifically, this is a method of leveraging PLMs by prepending task-specific prompts to the original input when feeding it into PLMs. One way to do this is with manually designed templates as task instructions~\citep{radford2019language, raffel2020exploring}; another one is to use continuous prompts that optimize a sequence of continuous task-specific vectors~\citep{lester-etal-2021-power, li-liang-2021-prefix}. More recently,~\citet{fu2022polyglot} have introduced an mT5-based framework to learn a unified semantic space blurring the boundaries of 6 NLP tasks with the prompting method, which we adopt in this work. 
Here, we investigate how a small PLM such as mt5 can be used in the multilingual multitask prompting framework, also to better understand how interrelated tasks can benefit from such a scheme.

Compared to very large models like GPT-3 \citep{brown2020}, smaller models  have the significant advantage of lower hardware requirements, making it easier to customize them quickly and cheaply for specific tasks, to implement modelling ideas iteratively, and for other researchers to reproduce experiments, too. Using a small PLM could however be very challenging when modelling more unrelated NLP tasks than those addressed in previous and in the current work, so this is something to bear in mind for future extensions.

\begin{table}[t]
\setlength{\tabcolsep}{8pt}
\resizebox{\linewidth}{!}{%
\centering
\begin{tabular}{lcccr}
\toprule
Form & Lang & Train & Valid & Test\\
\hline
\multirow{2}{*}{Hyperbole}
 & EN & 3,352 & 100 & 300\\
 & ZH & 3,760 & 600 & 1,000\\
 \hline
\multirow{4}{*}{Idiom}
 & EN & 18,676 & 1,470 & 200 [41/159]\\
 & DE & 14,952 & 1,670 & 200 [19/181]\\
 & ES & 12,238 & 1,706 & 199 [66/133]\\
 & IT & 15,804 & 1,732 & 200 [48/152]\\
 \hline
\multirow{4}{*}{Metaphor}
 & EN & 12,238 & 4,014 & 4,014\\
 & ES & 12,238 & 2,236 & 4,474\\
 & FA & 12,238 & 1,802 & 3,604\\
 & RU & 12,238 & 1,748 & 3,498\\

\bottomrule
\end{tabular}}
\caption{\label{tab:statistic}
Dataset Statistics. The label distribution is completely balanced (50\%/50\%), except for the idiom test sets, where the distribution is indicated in brackets as the proportion of literal/idiomatic next to the totals.
}
\end{table}

\begin{figure*}[!t]
\centering
\includegraphics[scale=.53]{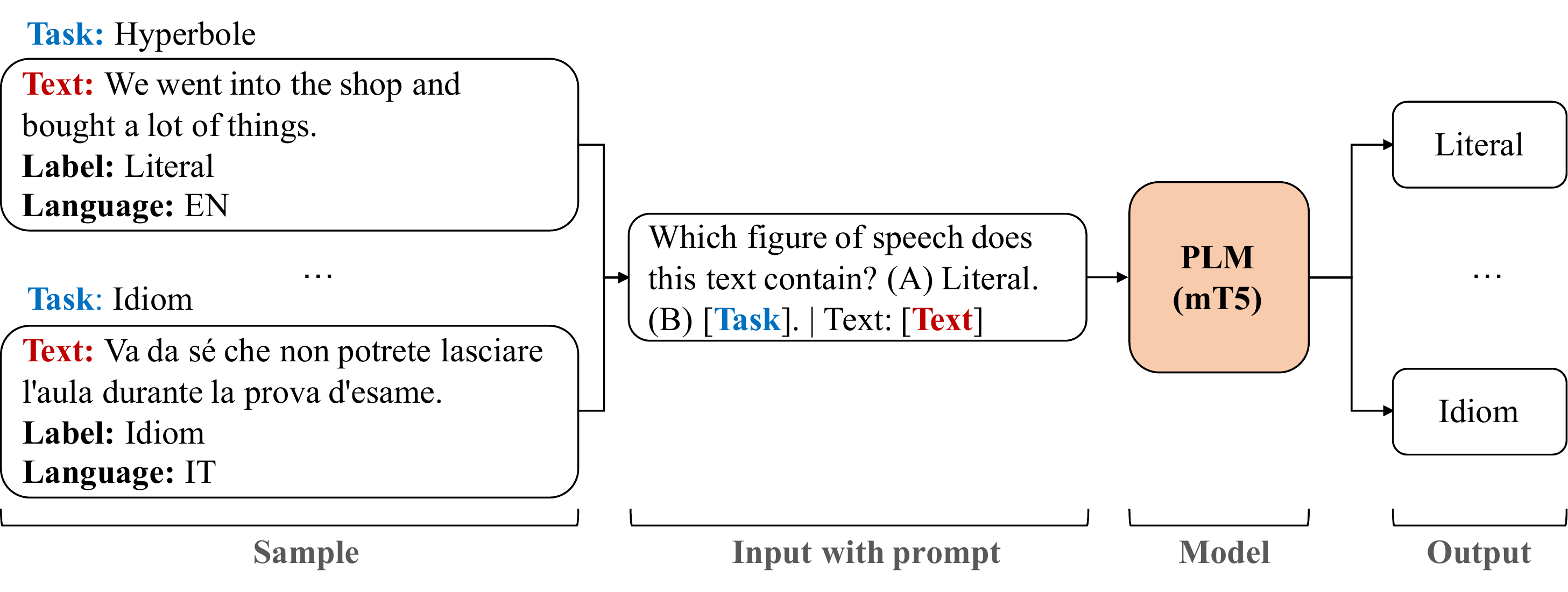}
\caption{
Overview for multilingual multi-figurative modelling based on prompt training. Given a detection task and the input text, we combine it with the predefined template, thus instructing the model to handle it.} 
\label{fig:approach}
\end{figure*}

\section{Tasks and Datasets}

\subsection{Task Formulation}

We focus on figurative language detection at sentence-level, which can be viewed as a binary classification task that requires identifying whether a given sentence is literal or non-literal (e.g.~idiomatic). To unify multiple figurative language detection tasks in different languages, we reformulate them as a text-to-text generation problem, where our model will generate the textual label for each given sentence.
For instance, given a sample $s$ from a detection task $T_{idiomatic} \in \emph{T}$, where $\emph{T} = \{T_{hyperbole}, T_{idiom}, T_{metaphor}\}$ is the task set we consider, the model aims to output the text label $y \in \{\textrm{Literal}, \textrm{Idiomatic}\}$.


\subsection{Datasets}

We use five existing figurative language datasets for our experiments, which cover three figures of speech and seven languages. 
Table~\ref{tab:statistic} shows the dataset statistics for the various languages in each figure of speech.

\paragraph{Hyperbole}
HYPO~\citep{troiano-etal-2018-computational} is an English dataset containing 709 hyperbolic sentences with their corresponding non-hyperbolic versions. HYPO-Red~\citep{tian-etal-2021-hypogen-hyperbole} is another dataset that includes literal and hyperbolic texts. We combine these two datasets for the English hyperbole detection task. HYPO-cn~\citep{kong-etal-2020-identifying} is a Chinese hyperbole detection dataset. Since both English and Chinese hyperbole datasets are rather 
small compared to the sizes of the training datasets for the other figures of speech, we upsample them by random instance replication 
obtaining training sets of 10,000 samples.

\paragraph{Idiom} 
ID10M~\citep{tedeschi-etal-2022-id10m} is a multilingual idiom dataset, containing automatically-created (silver) training and validation 
data in 10 languages and manually-created (gold) test sets in 4 languages: English, German, Italian, Spanish. This dataset is designed for word-level idiom detection; we convert it to sentence-level labels and use the four languages with gold data.

\paragraph{Metaphor}
LCC~\citep{mohler-etal-2016-introducing} is a multilingual metaphor dataset derived from web-crawled data in four languages: English, Spanish, Russian, and Farsi. It provides metaphoricity ratings for within-sentence word pairs on a four-point scale, including 0 as no, 1 as weak, 2 as Conventional, and 3 as clear metaphor. We use the data preprocessed by~\citet{aghazadeh-etal-2022-metaphors}.

\section{Multilingual Multi-Figurative Model}

We propose a multitask and multilingual framework based on template-based prompt learning for figurative language detection. 

\subsection{Multitask Prompt Training}

We use mT5~\citep{xue-etal-2021-mt5} as our backbone and jointly model multiple detection tasks, with the ultimate goal of having one single model that can handle the detection of multiple figures of speech in multiple languages. 
The overall framework is illustrated in Figure~\ref{fig:approach}. Given a sample $x$ from the $t^{th}$ task $T_{t}$, it is first combined with the predefined prompt template $p_{t}$ and then fed into model $M$, which is expected to produce the label $y$: $M (x, p_{t}) = y^{'}$. 

We minimize the negative log-likelihood of the sequences of the model’s outputs, the loss function being formulated as:

\begin{equation}\label{eq:loss}
    L_{\theta} = -\sum \log (\boldsymbol{y}\mid \boldsymbol{x}; \theta)
\end{equation}

\noindent where $\theta$ are the parameters of mT5. $\boldsymbol{x}$ and $\boldsymbol{y}$ represents the sequences of the given sentence $x$ and its 
text label $y$, respectively. We use the multilingual and multi-figurative samples from dataset $T$ to fine-tune mT5, adapting it to the figurative language detection tasks. 

We design the prompt templates based on our intuition of how we would ask a human annotator to complete the figurative language detection task. In our main framework, we use a cross-lingual template setting whose templates for all tasks are in English. We will assess the impact of different prompts settings, including template and language (see Sec~\ref{sub:analysis}).

\subsection{Generalization}

We investigate the generalization ability of our proposed framework in cross-figurative and cross-lingual scenarios, where the training and test data come from different figure-of-speech or different languages.


\paragraph{Cross-Figurative Knowledge}

Inspired by~\citet{lai-nissim-2022-multi}, we evaluate our framework in terms of cross-figurative knowledge transfer. The hypothesis is that different figures of speech might share some figurative features, and that a text may contain different figures of speech simultaneously, possibly triggered by different textual portions, so that a single framework might warrant a large knowledge gain through transfer from one  figure of speech to another.

Using a multitask framework to jointly model multiple figurative language detection tasks, the cross-figurative generalization ability is expected to improve performance across different tasks compared to single figurative language modelling.

\paragraph{Cross-Lingual Knowledge}
Multilingual PLMs are pre-trained on 
texts from multiple languages
in a self-supervised way, which enables different languages to be represented in 
a single space. Therefore, words and phrases that are similar across languages will be close to each other. We extend and evaluate the cross-lingual generalization in metaphor carried out by~\citet{aghazadeh-etal-2022-metaphors} to a setting with multiple figures of speech. The hypothesis is that if the knowledge of figurative language is transferable across languages, then the model $M_{l_{m}}$ would be able to have a good generalization in language $l_{n}$ based on what it has learned in language $l_{m}$: $M_{l_{m}} (x, p_{t}) = y^{'}$, for ($x$, $y$) $\in$  $T_{t}$ in language $l_{n}$. 
Furthermore, cross-lingual knowledge transfer can further improve model performance when doing multilingual modelling.

However,
cultural differences often have a great influence on the usage of figurative language. Idioms are culture-/language-specific, for example, with established meanings over a long period of usage in a specific cultural background~\citep{geoffrey1994}. Therefore, we expect that the model will have different performances in cross-lingual generalization for different figures of speech depending on how culturally-related the languages involved are.

\section{Experiments}

\subsection{Setup}

We use mT5-base 
(580M parameters) to evaluate our  framework. All experiments are implemented atop Transformers~\citep{wolf-etal-2020-transformers}.  
We train our models with batch size 32, using the Adam optimiser~\citep{kingma2017adam} with a polynomial learning rate decay. We set a linear warmup of 1,000 steps for a maximum learning rate of 1e-4 and a maximum decay of 10,000 steps for a minimum learning rate of 5e-5. We evaluate checkpoints every 1,000 steps, and use early stopping (patience 5) if validation performance does not improve. Following~\citet{aghazadeh-etal-2022-metaphors}, we report performance as detection accuracy for all experiments. In Section~\ref{sub:analysis}, we include an additional analysis for the unbalanced idiom datasets.

\subsection{Model Settings}

Since we take the first step towards the joint modelling for multilingual multi-figurative language detection, we conduct extensive experiments with different architectures and settings, leading to five sets of models. Additionally, we obtain zero-shot results in non-English languages by utilizing English-only variants of the same sets of models. 


\begin{itemize}[leftmargin=*]
\itemsep 0in
\item \textbf{Baseline} Following \citet{tayyar-madabushi-etal-2022-semeval}, we train a binary detection classifier for each figure of speech and language by fine-tuning multilingual BERT~\citep{devlin-etal-2019-bert}. Our work is similar to one previous work on sentence-level metaphor detection, which is carried out by \citet{aghazadeh-etal-2022-metaphors} in a multilingual setting. However, they assume that the phrase in the sentence to be classified as metaphoric or not is already known in advance, while our models do not use such information. Therefore, we do not consider it as a baseline model.

\item \textbf{Vanilla mT5} Similar to mBERT, we fine-tune mT5 on specific figures of speech for each language but in a seq2seq fashion.

\item \textbf{Prompt mT5} We fine-tune mT5 with the prompt template in a seq2seq way for each figure of speech in each language with the aforementioned sets of models.

\item \textbf{+ multitask} These are multilingual multitask models. We fine-tune mT5-based models with their corresponding single-task training methods using all data from $\emph{T}$.

\item \textbf{Zero-shot with EN model} Based on the above models, but we train them on English data only and test them on non-English languages.

\end{itemize}

\begin{table*}[!t]
\setlength{\tabcolsep}{10pt}
\resizebox{\linewidth}{!}{%
\centering
\begin{tabular}{lcccccccccccc}
\toprule
\multirow{2}{*}{\textbf{Model}} & \multicolumn{2}{c}{\textbf{Hyperbole}} & & \multicolumn{4}{c}{\textbf{Idiom}} & & \multicolumn{4}{c}{\textbf{Metaphor}}\\
\cline{2-3} \cline{5-8} \cline{10-13}
 & EN & ZH & & EN & DE & ES & IT & & EN & ES & FA & RU\\
 \midrule
 \multicolumn{13}{l}{Main results}\\
\hline
 Baseline           & 72.33 & 80.40 &  & 79.00 & 72.50 & 66.33 & 70.50 &  & 81.37 & 80.11 & 74.83 & 79.93\\
 Vanilla mT5     & 72.67 & 71.40 &  & 79.50 & 74.50 & 64.82 & \textbf{76.00} &  & 82.64 & 82.32 & 77.33 & 82.25\\
 \ \ + multitask & 72.67 & 81.40 &  & 62.00 & 74.50 & 56.78 & 72.00 &  & 81.86 & 81.20 & 77.61 & \underline{\textbf{83.76}}\\
 Prompt mT5      & 81.00 & 81.60 &  & 79.50 & 75.00 & \textbf{68.34} & 75.00 &  & \underline{\textbf{83.43}} & 82.66 & 76.64 & 83.39\\
 \ \ + multitask & \textbf{82.00} & \underline{\textbf{82.60}} &  & \underline{\textbf{86.00}} & \underline{\textbf{79.00}} & 67.84 & \textbf{76.00} &  & 83.06 & \underline{\textbf{83.10}} & \underline{\textbf{78.14}} & 83.16\\
\midrule
\multicolumn{13}{l}{(Zero-shot) with EN model}\\
\hline
 Baseline           & 72.33 & 69.60 &  & 79.00 & 62.00 & 61.81 & 60.00 &  & 81.37 & 71.70 & 61.29 & 69.01\\
 Vanilla mT5     & 72.67 & 70.20 &  & 79.50 & 53.00 & 64.32 & 70.50 &  & 82.64 & 75.10 & 68.70 & 76.10\\
 \ \ + multitask & 65.67 & 64.90 &  & 72.50 & 52.50 & 37.69 & 63.50 &  & 82.41 & 71.86 & 66.84 & 73.61\\
 Prompt mT5      & 81.00 & 74.00 &  & 79.50 & 59.00 & \underline{\textbf{69.85}} & 76.50 &  & \underline{\textbf{83.43}} & \textbf{75.95} & \textbf{70.17} & \textbf{76.39}\\
 \ \ + multitask & \underline{\textbf{82.33}} & \textbf{76.10} &  & \textbf{81.50} & \textbf{65.60} & 66.83 & \underline{\textbf{79.50}} &  & 81.27 & 74.99 & 68.70 & 75.93\\
\bottomrule
\end{tabular}}
\caption{\label{tab:main-results}
Results (accuracy) for multilingual multi-figurative language detection, covering three figures of speech and seven languages. Notes: (i) we include results on English tasks for the block of zero-shot modelling with the EN model for comparison with those included in the main results; (ii) bold numbers indicate the best systems for each block, and underlined numbers indicate the best score for each language.
}
\end{table*}

\subsection{Results}

Table~\ref{tab:main-results} reports results on three figurative language detection tasks in seven languages.

\paragraph{Main Results} 
We see that \texttt{Vanila mT5} performs better than \texttt{mBERT} on most tasks, except ZH hyperbole and ES idiom. When \texttt{Vanila mT5} is used for multitask training, unsurprisingly, its performance drops in many tasks. One straight reason is that it is challenging to model multiple tasks at once. The other possible reason is that a text may contain features of multiple figures of speech at the same time, but there is not enough evidence to guide the model to perform a specific task. 
In other words, the model may correctly predict the figurative form for a given text, but it does not match the label of the target task. 

When looking at \texttt{Prompt mT5}, we see that the model with prompt training brings improvement for most tasks compared to \texttt{Vanila mT5}. This shows the effectiveness of the prompt, which instructs the model to perform the target task. \texttt{Prompt mT5} with multi-task training has the best performances on most tasks: (i) it shows a steady 
improvement in hyperbole detection; (ii) in idiom detection performances are boosted for EN, DE, and IT though the ES score is lower compared to \texttt{Prompt mT5}; (iii) for metaphor detection it achieves the highest accuracy in ES and FA but slightly underperforms in EN and RU compared to \texttt{Prompt mT5}.

\paragraph{Zero-Shot} For zero-shot results on non-EN languages using EN models, we see similar trends to the main results (see Table~\ref{tab:main-results}, second block). \texttt{Vanilla mT5} has overall better performances than its multitask counterpart and \texttt{mBERT}. We observe that \texttt{Prompt mT5}-based models 
have a clear edge in this setting, with the highest accuracy for all tasks and languages obtained by one of them. EN models yield the highest accuracy scores in EN hyperbole and metaphor detection, and even in idiom detection of ES and IT with zero-shot. The main reason for this is most likely that the idiom training and validation data is created automatically, leading to a non-test set of inferior quality and reduced performance on the test set compared to the validation set (see Sec~\ref{sub:analysis}). 
Overall, a zero-shot approach for figurative language detection when lacking high-quality resources in the target language seems a highly reliable strategy.

\subsection{Analysis and Dissussion}
\label{sub:analysis}

\begin{table}[t]
\setlength{\tabcolsep}{6pt}
\resizebox{\linewidth}{!}{%
\centering
\begin{tabular}{l|ccc|ccc}
\toprule
Form & Lang & Valid & Test & Lang & Valid & Test\\
\hline
Hyperbole  
  & EN & 87.00 & 82.00 & ZH & 83.00 & 82.60\\
  \midrule
\multirow{2}{*}{Idiom}     
  & EN & 70.07 & 86.00 & DE & 97.01 & 79.00\\
  & ES & 91.68 & 67.84 & IT & 94.40 & 76.00\\
 \midrule
\multirow{2}{*}{Metaphor}
  & EN & 83.06 & 83.06 & ES & 83.54 & 83.10\\
  & FA & 78.30 & 78.14 & RU & 82.78 & 83.16\\
\bottomrule
\end{tabular}}
\caption{\label{tab:valid-test-results}
Results (accuracy) of our main model (\texttt{Prompt mT5} + multitask) on validation and test sets.
}
\end{table}

\begin{table}[t]
\setlength{\tabcolsep}{11pt}
\resizebox{\linewidth}{!}{%
\centering
\begin{tabular}{ccccc}
\toprule
\multirow{2}{*}{Lang} & \multicolumn{2}{c}{Valid} & \multicolumn{2}{c}{Test}\\
\cline{2-5}
& Literal & Idiomatic & Literal & Idiomatic\\
\midrule
EN & 34.83 & 49.12 & 48.78 & 62.26\\
DE & 2.16  & 97.61 & 63.16 & 65.75\\
ES & 18.17 & 97.19 & 13.64 & 30.83\\
IT & 5.88  & 98.28 & 85.42 & 68.42\\
\bottomrule
\end{tabular}}
\caption{\label{tab:idiom-statistic}
The ratio (\%) of idiomatic expressions contained in sentences of valid/test sets that appear in idiomatic sentences of training sets.
}
\end{table}

\begin{figure*}[!t]
\centering
\includegraphics[scale=.39]{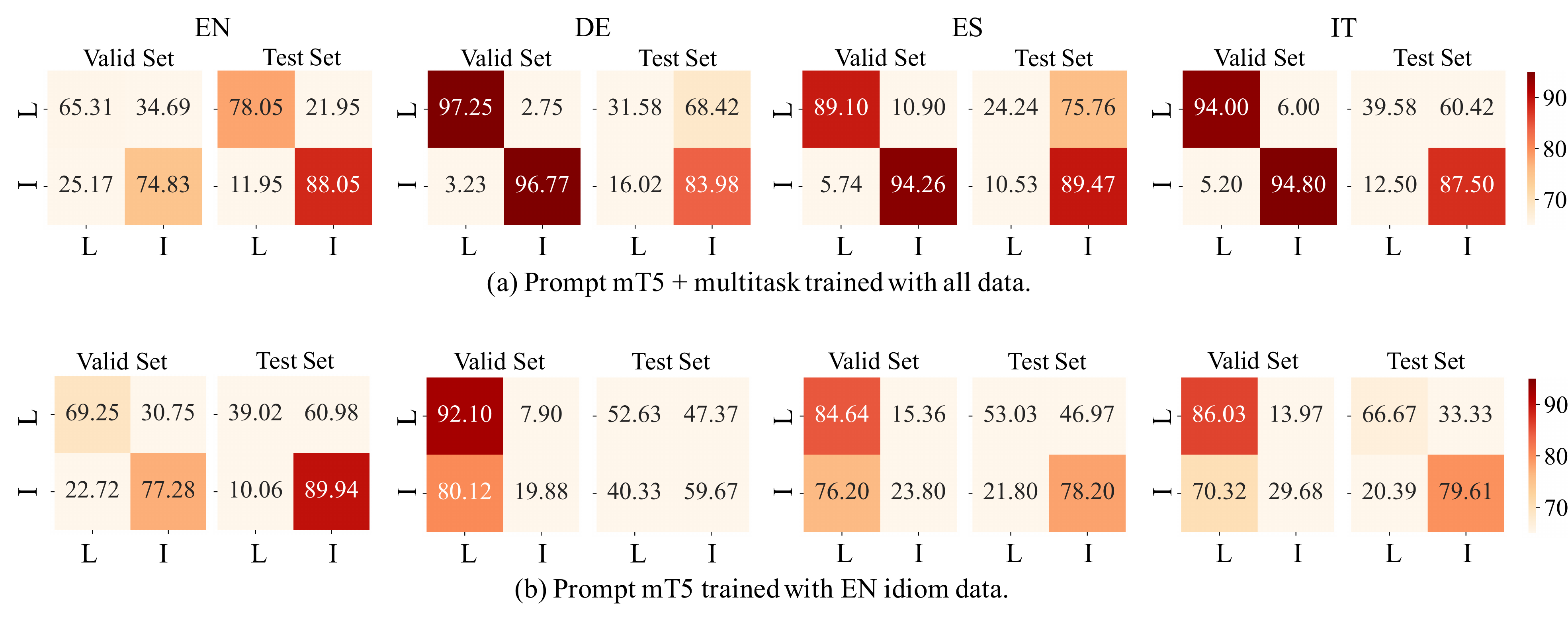}
\caption{Confusion matrix of main model and zero-shot on idiom detection, where predictions (X-axis) are compared to the corresponding ground truth labels (Y-axis). L~=~Literal and I~=~Idiomatic.} 
\label{fig:idiom-confusion}
\end{figure*}

\paragraph{Error Analysis.}
Table~\ref{tab:valid-test-results} presents the results of our main model (\texttt{Prompt mT5} + multitask) on validation and test sets. The performances on the test sets are comparable to the validation sets for hyperbole and metaphor, while the idiom task stands out: for EN idiom detection, test accuracy is higher than validation accuracy, while we observe the opposite in languages other than English, with validation scores above 90\% and test scores below 80\%.

To analyze this behaviour of the idiom task, in Table~\ref{tab:idiom-statistic} we report the ratio of idiomatic expressions contained in sentences of the validation/test sets that also appear in idiomatic sentences of the training sets. The distribution of the EN data is relatively balanced for both validation and test. For other languages, most of the expressions in the idiomatic sentences of the validation set are already present in the training set, but this is not the case for the literal sentences. Regarding test sets, the ratio of DE and IT are very high for both literal and idiomatic sentences, but very low for ES, which poses a significant challenge to the model.

We also group both the predictions and the labels to produce the confusion matrices for the main and EN models in Figure~\ref{fig:idiom-confusion}. For the EN task, we see that scores on the main diagonal are higher than those on the secondary diagonal, except for the EN model in the test set (39.02 vs 60.98). This is commonly observed in binary classification experiments. We have different observations on in-language training and zero-shot in other languages: (i) from (a), we see that the in-language model performs very well on the validation set for literal and idiomatic sentences, while it overpredicts literal sentences on the test set; (ii) in contrast, EN models in (b) perform better in the test sets and they overpredict idiomatic sentences on the validation set.

Generally, a sentence might not be idiomatic as a whole although it contains idioms. Such sentences will be labelled as non-literal in the automatic dataset creation. Based on the above observations, we see that the distribution of the automatically created training and validation data is quite different from the manually created test set, and the quality of the former is much lower than that of the latter. The nature of training data actually affects the stability of the model on different tasks. For instance, this even leads to better performances using EN models (zero-shot) on ES and IT than using in-language-trained models (see Table~\ref{tab:main-results}).

\begin{table}[t]
\setlength{\tabcolsep}{8pt}
\resizebox{\linewidth}{!}{%
\centering
\begin{tabular}{lcccc}
\toprule
Model & Lang & Hyperbole & Idiom & Metaphor\\
\hline
Prompt mT5      & \multirow{2}{*}{EN} & 81.00 & 79.50 & 83.43\\
\ \ + multitask & & 82.33 & 81.50 & 81.27\\
\hline
Prompt mT5      & \multirow{2}{*}{ES} & - & 68.34 & 82.66\\
\ \ + multitask & & - & 70.35 & 82.14\\
\bottomrule
\end{tabular}}
\caption{\label{tab:cf-results}
Results (accuracy) for single figurative form modelling and cross-figurative modelling in English and Spanish.
}
\end{table}

\paragraph{Cross-Figurative Knowledge Transfer}

To further investigate cross-figurative knowledge transfer, we sample different figures of speech for two languages 
from our dataset and compare single- to multi-figurative language modelling. 
Table~\ref{tab:cf-results} shows the results for EN and ES.
For EN, compared to single figurative form models, we see that multitask modelling yields further improvements in hyperbole and idiom but hurts metaphor. Similarly, when combining information on both idioms and metaphors for ES, extra information about idioms hurts metaphor detection slightly, while extra information about metaphors helps idioms. We suggest two main reasons for these observations: (i) performance improvements in hyperbole and idioms are enhanced by the transfer of knowledge from metaphors; (ii) The low-quality idiom training data, as discussed earlier in this section, negatively impacts the accuracy of metaphor detection. While incorporating information from hyperbole data could potentially be beneficial, the limited amount of such data might not be enough to bring any benefit.

\begin{figure*}[!t]
\centering
\includegraphics[scale=.57]{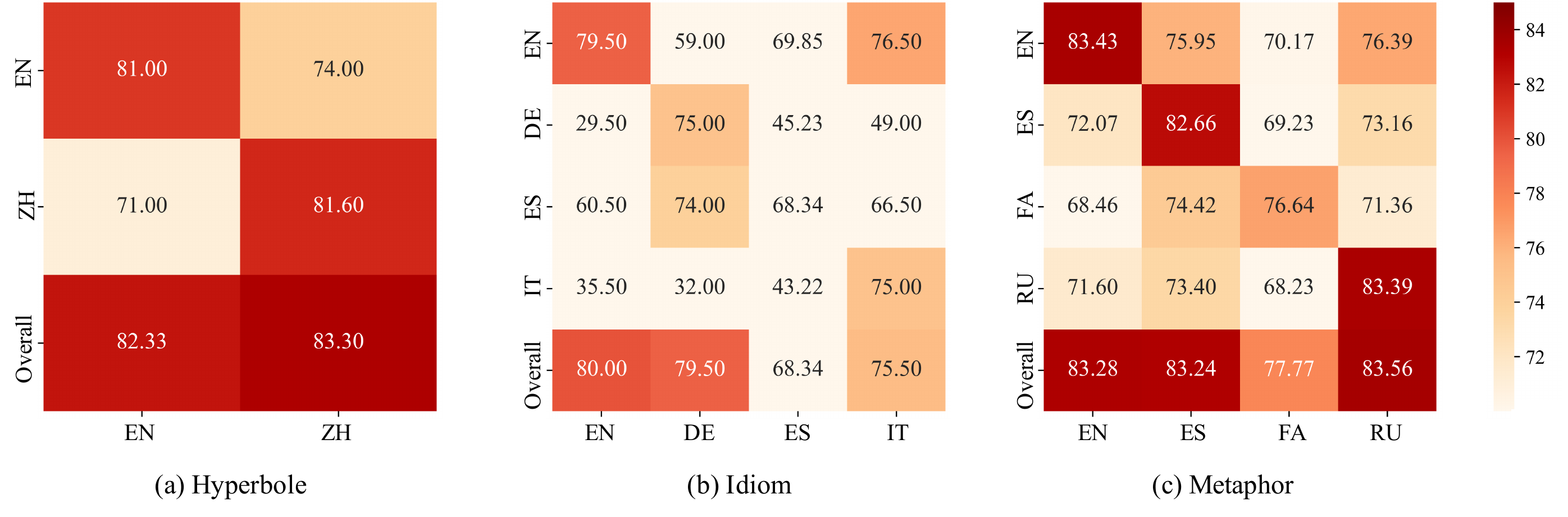}
\caption{Cross-lingual detection results (accuracy). 
The y-axis shows the language used for training (overall stands for a model trained on all the languages) while the x-axis indicates the language in which the model is tested on.} 
\label{fig:cross-lingual}
\end{figure*}

\paragraph{Cross-Lingual Knowledge Transfer}

We use the model trained in one language to run the zero-shot experiments on the other languages and model all languages jointly for each figure of speech. Figure~\ref{fig:cross-lingual} shows the results for cross-lingual experiments. Zero-shot has moderate detection accuracy on hyperbole and metaphor with scores greater than 68\% for all languages, confirming that figurative knowledge is transferable across languages. In idiom detection, it is unsurprising to see that zero-shot performs poorly, e.g.\ the accuracy of the DE model on EN idiom detection is only 29.5\% considering that cultural specificities of idioms might hamper
cross-lingual generalization more than for other figures of speech. Still, multilingual modelling brings performance improvements on most tasks in different languages, including idioms. 
Overall, figurative language detection can benefit from multilingual modelling, and the zero-shot technique can be used for hyperbole and metaphor detection when lacking resources in the target language, but not, in most cases, for idiom detection.

\begin{table*}[!t]
\setlength{\tabcolsep}{5pt}
\resizebox{\linewidth}{!}{%
\centering
\begin{tabular}{llll}
\toprule
Task Lang & Prompt Lang & \# & Prompt Template\\
\hline
\multirow{4}{*}{Italian} & \multirow{3}{*}{English} & $A$ & Which figure of speech does this text contain? (A) Literal. (B) [\texttt{TASK}]. | Text: [\texttt{IT-text}] \\
 & & $B$ & Is there a(n) [\texttt{TASK}] in this text? | Text: [\texttt{IT-text}]\\
 & & $C$ & Does this text contain a(n) [\texttt{TASK}]? | Text: [\texttt{IT-text}]\\
 \cline{2-4}
 & Italian & $D$ & Quale figura retorica contiene questo testo? (A) Letterale. (B) [\texttt{TASK}]. | Testo: [\texttt{IT-text}]\\
\bottomrule
\end{tabular}}
\caption{\label{tab:prompts}
Examples of different prompt templates. \texttt{TASK} and \texttt{IT-text} represent the placeholders for figure of speech (e.g.\ idiom) and the text in Italian, respectively.
}
\end{table*}

\begin{figure}[t]
    \begin{minipage}[t]{0.7\linewidth}
    \centering
    \subfigure[Results between templates $A$ and $B$.]{
      \includegraphics[scale=0.52]{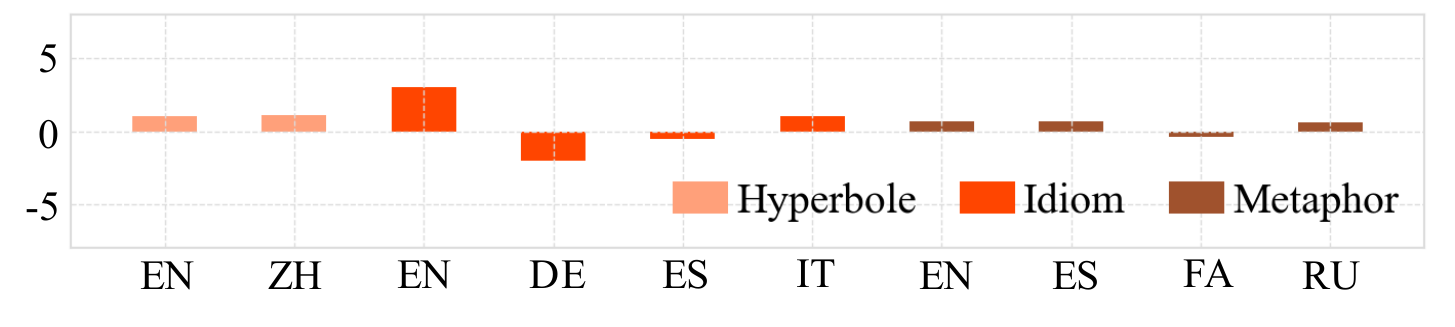}
      \label{fig:diff0}
    }
    \end{minipage}

    \vspace{1.0mm}
    \begin{minipage}[t]{0.7\linewidth}
    \centering
    \subfigure[Results between templates $A$ and $C$.]{
      \includegraphics[scale=0.52]{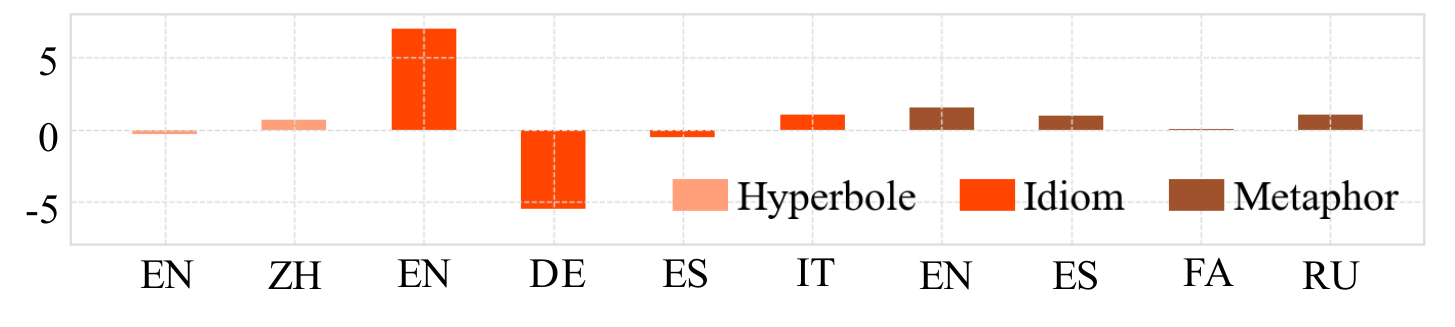}
      \label{fig:diff1}
    }
    \end{minipage}

    \vspace{1.0mm}
    \begin{minipage}[t]{0.7\linewidth}
    \centering
    \subfigure[Results between cross-lingual and in-lingual templates ($A$ VS $D$).]{
      \includegraphics[scale=0.52]{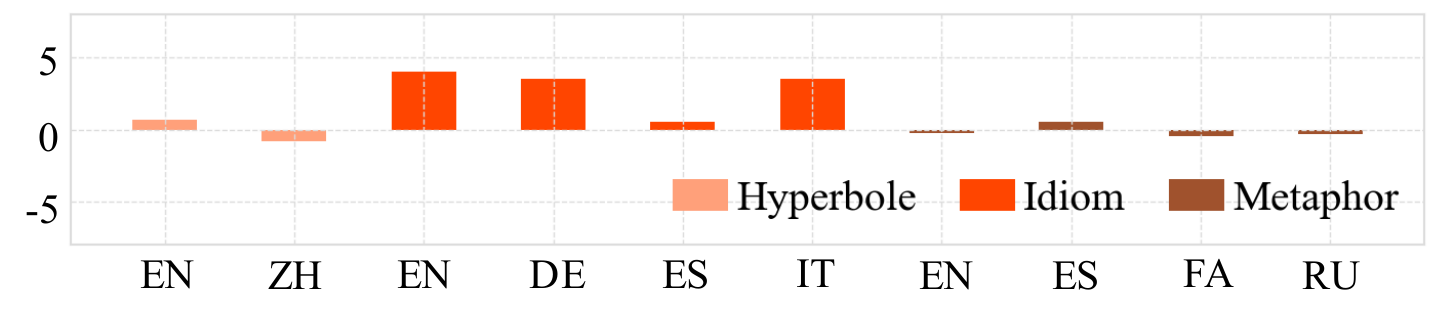}
      \label{fig:diff2}
    }
    \end{minipage}
    
    \caption{Relative performance differences between different prompt templates.}
    \label{fig:diff}
\end{figure}

\paragraph{Impact of Prompt}

Although prompt learning has been shown an effective method for many NLP tasks, it usually requires extensive prompt engineering on template design as it is sensitive to different tasks. Following~\citet{fu2022polyglot}, we assess the prompt effect with different templates and languages. 
In Table~\ref{tab:prompts}, we show a set of cross-lingual and in-lingual prompt templates. In the cross-lingual prompt setting, all templates are written in English, here we experiment with two other templates ($B$ and $C$) besides the one used in our main results ($A$). 
In the in-lingual prompt setting, instead, the language of the template is consistent with the task language. Template~$D$ in Table~\ref{tab:prompts}, for example, is an in-lingual prompt translated from~$A$ and used for Italian tasks.

Figure~\ref{fig:diff} shows the relative performance differences, where we subtract the performance of the model with other prompts from the model with prompt $A$. In the cross-lingual setting, we see that the performances of models with different prompt templates are very close, with an accuracy difference of less than 5 percentage points on all tasks except for EN and DE 
idiom using template $A$ and $C$. Interestingly, the English template does not hurt performances in other languages (Figure~\ref{fig:diff2}). These results suggest that a model based on prompt learning for multilingual multi-figurative language detection is not particularly sensitive to different templates.

\section{Conclusions}

We introduced a multilingual multi-figurative language understanding benchmark that focuses on sentence-level figurative language detection, involving three common figures of speech and seven languages. Based on prompt learning, we proposed a framework to unify the interrelated detection tasks across multiple figures of speech and languages using a PLM, while having no task- or language-specific modules. We further analyzed the generalization of the model across 
different
figures of speech and languages. 

Our  unified model benefits from cross-lingual and cross-figurative knowledge transfer in sentence-level detection. It is natural to explore fine-grained detection at the word-level in future work, as well as language generation in multilingual and multi-figurative scenarios. 
This approach can also serve as a blueprint for the joint modelling of other interrelated tasks.

\section{Limitations and Impact}

While introducing a framework which deals with multiple languages and multiple figures of speech, this work is still only dealing with three figures of speech and seven languages. Many more phenomena and languages can still bring substantial challenges and insights if considered (once the data availability bottleneck is addressed). Also, we deal with figurative language as labelled at the sentence level, but the word level is also not only interesting but important for broader natural language understanding and could yield different insights than those observed in the present work. 

We only mention in passing the influence that different cultural contexts have on figurative usages, and we make some observations on idioms, but this aspect would require a much bigger unpacking. We actually believe that (failure) of cross-lingual computational models can be an excellent diagnostic tool towards a finer-grained analysis of the interplay between culture(s) and figurative language. 

We propose a successful method based on prompt learning and present experiments using a specific pre-trained model. Choosing different (and possibly larger) models and investigating even more than what we already do in this paper the influence of specific prompts would also be necessary to further generalise the efficacy of our approach. 

Finally, as with most language technology, the limitations of our approach, also in terms of accuracy (especially for some phenomena and some languages), could lead to substantial inaccuracies which could be propagated in further processing. Considering that figures of speech are associated with emotional language, a word of warning is necessary regarding the direct deployment of our models. We do hope that writing about risks explicitly and also raising awareness of this possibility in the general public are ways to contain the effects of potential harmful consquences. We are open to any discussion and suggestions to minimise such risks.

\section*{Acknowledgments}

This work was partly funded by the China Scholarship Council (CSC). The anonymous reviewers of ACL 2023 provided us with useful comments which contributed to improving this paper and its presentation, so we're grateful to them. We would also like to thank the Center for Information Technology of the University of Groningen for their support and for providing access to the Peregrine high performance computing cluster.

\bibliography{anthology,custom}
\bibliographystyle{acl_natbib}




\end{document}